%% file: 0_abstract.tex
\def\year{2022}\relax
\documentclass[letterpaper]{article} % DO NOT CHANGE THIS
\usepackage{aaai22}  % DO NOT CHANGE THIS
\usepackage{times}  % DO NOT CHANGE THIS
\usepackage{helvet}  % DO NOT CHANGE THIS
\usepackage{courier}  % DO NOT CHANGE THIS
\usepackage[hyphens]{url}  % DO NOT CHANGE THIS
\usepackage{graphicx} % DO NOT CHANGE THIS
\usepackage{tabularx}
\def\UrlFont{\rm}  % DO NOT CHANGE THIS
\usepackage{natbib}  % DO NOT CHANGE THIS AND DO NOT ADD ANY OPTIONS TO IT
\usepackage{caption} % DO NOT CHANGE THIS AND DO NOT ADD ANY OPTIONS TO IT
\DeclareCaptionStyle{ruled}{labelfont=normalfont,labelsep=colon,strut=off} % DO NOT CHANGE THIS
\usepackage{algorithm}
\usepackage{algorithmic}
\usepackage{multirow}
\usepackage{subfigure}
\usepackage{hyperref}

\usepackage{amssymb}
\usepackage{lscape}
\hypersetup{colorlinks,linkcolor={blue},citecolor={blue},urlcolor={blue}}
\usepackage{amsmath}

\usepackage{pifont}% http://ctan.org/pkg/pifont
\newcommand{\cmark}{\ding{51}}%
\newcommand{\xmark}{\ding{55}}%

\usepackage{newfloat}
\usepackage{listings}
\floatstyle{ruled}
\newfloat{listing}{tb}{lst}{}
\floatname{listing}{Listing}

\usepackage{comment}
\title{Relative Switching Point based Dynamic Positional Representation for Code-Mixed Sentiment Analysis}
\title{PESTO: Switching Point based Dynamic and Relative Positional Encoding for Code-Mixed Languages}
\author {
    Mohsin Ali\textsuperscript{\rm 1} \quad
    Kandukuri Sai Teja \textsuperscript{\rm 1} \quad
    Sumanth Manduru\textsuperscript{\rm 1} \\
    Parth Patwa\textsuperscript{\rm 2} \quad
    Amitava Das \textsuperscript{\rm 3, 4}
}
\affiliations {
    IIIT Sri City, India\textsuperscript{\rm 1} \quad
    UCLA, USA\textsuperscript{\rm 2} \quad
    Wipro AI Labs, India\textsuperscript{\rm 3} \quad
    AI Institute, University of South Carolina, USA\textsuperscript{\rm 4}
    \\
    \{mohsinali.m18, saiteja.k18, sumanth.m15\}@iiits.in,  parthpatwa@ucla.edu, amitava.das2@wipro.com
}
\begin{document}

\maketitle

\begin{abstract}

% In this paper, we study the code-mixing problem. We collect and train embeddings on code-mixed data using various word embedding methods. Further, we propose a novel variable length multi head attention with relative switching-point based dynamic positional representation for sentiment analysis task on the SemEval-2020 data and achieve state-of-the0art results of 75.56\% F1 score.

NLP applications for code-mixed (CM) or mix-lingual text have gained a significant momentum recently, the main reason being the prevalence of language mixing in social media communications in multi-lingual societies like India, Mexico, Europe, parts of USA etc. Word embeddings are basic building blocks of any NLP system today, yet, word embedding for CM languages is an unexplored territory. %On the other hand, it is known that positional info and/or sequence order is an important aspect in NLP. %Positional aspects are again in discussion as people started realising that transformer based models over-simplified positional info using $sin/cos$ functions. 
%There are several challenges for CM word embedding - spelling variations, word play, different languages etc. 
The major bottleneck for CM word embeddings is switching points, where the language switches. These locations lack in contextually and statistical systems fail to model this phenomena due to high variance in the seen examples. In this paper we present our initial observations on applying switching point based positional encoding techniques for CM language, specifically \textit{Hinglish} (Hindi - English). Results are only marginally better than SOTA, but it is evident that positional encoding could be an effective way to train position sensitive language models for CM text.

%Code mixing is a common phenomenon in social media now a days. In this paper, we explore the problem of sentiment analysis of code mixed \textit{Hinglish} text. We have collected Hinglish twitter data and trained various embedding models on the data and evaluated these embeddings on similarity files for intrinsic evaluation. One of the biggest challenges for sentiment analysis on code mixed data is switching points. Therefore, to address the problem of sentiment analysis for code-mixed data, we have designed a novel neural architecture that is sensitive towards switching points. Our network is a variable length multi head attention model with relative and switching point based dynamic positional representations. We have achieved state of the art result of weighted F1 score of 75.56\% on semeval-2020 data.

\end{abstract}

\input{1_introduction}

\input{3_positional_encoding}
\input{4_experiment_result}

\input{5_conclusion}

%\bibliographystyle{aaai22}
\fontsize{9pt}{10pt}\selectfont
\bibliography{aaai22.bib}

\end{document}

%% file: 1_introduction.tex
%\vspace{-6mm}
\section{Switching Points: The Bottleneck}

Switching Points (SPs) are the positions in CM text, where the language switches. %For code-mixed languages, consisting of a pair of languages, there can be two types of switching points. For code-mixed languages consisting of two languages \textit{L1} and \textit{L2}, SP occurs when the language changes from L1 to L2 or L2 to L1.
Consider the text - \textit{aap}\textsubscript{HI} \textit{se}\textsubscript{HI} \textit{request}\textsubscript{EN} \textit{hain}\textsubscript{HI}  (request you to). Here, when the language switches from \textit{Hindi}  to \textit{English} (\textit{se}\textsubscript{HI} \textit{request}\textsubscript{EN}) a \textbf{HI-EN} (HIndi-ENglish) SP occurs. Correspondingly, a \textbf{EN-HI} SP occurs at \textit{request}\textsubscript{EN} \textit{hain}\textsubscript{HI}. %In the context of modeling CM languages, switching points can be considered as ordinary bigrams that occur with other monolingual bigrams in a corpus. It is easy to infer that particular SP bigrams will be relatively rare in a given corpus. Hence, such sparse occurrences of switching point bigrams makes it difficult for any Language Model (LM) to learn their probabilities and context. % LMs have two major forms - word embeddings (can consider as LM or can be argued to be motivated from LM), and LMs for sentence levels -  big LMs like BERT, GPT,  etc.
In this work we to look at sentiment analysis of CM languages, specifically \textit{Hinglish} through the lens of language modeling. We propose PESTO -  a switching point based dynamic and relative positional encoding. PESTO learns to emphasis on switching points in CM text. % Relativity here is relative distances of words from SPs. 
Our model marginally outperforms the SOTA.
%We also introduce the idea of switching-point based relative positional encoding.

\begin{comment}
There has been a significant increase in the usage of Indian languages in informal multilingual communications such as twitter, Whatsapp messages, blogs, etc over the past few years. Code mixing, which means mixing two languages together to form understandable and structured phrases, is a norm in bi or multilingual regions such as Asia, Europe etc. To convey the emotions as desired, people use their native language along with English words for easier understanding. The following phrase is an example of code mixing of Hindi with English. 

\begin{quote}
\centering
\textit{Aye}\textsubscript{HI} \textit{aur}\textsubscript{HI} \textit{enjoy}\textsubscript{HI}
\textit{kare}\textsubscript{HI}
\\
\textbf{English Translation: } 
\end{quote}

With the growth of availability of code mixed data in social media, chats, blogs etc, there has been a growing interest towards analysing the sentiment posted in code mixed text. Since many low resource languages like Hindi, Spanish and Bengali are used along with English, traditional sentiment analysis models perform poorly over this data. Extracting the sentiments from code mixed data is challenging for many reasons such as spelling variations, grammar mismatch, no word order, switching points((junctions in the text where there is a switch in language), etc. \\

\end{comment}

%% file: 3_positional_encoding.tex
\section{Background - Dataset and  Positional Encoding }

\textbf{Data and SOTA}: 
The SentiMix task @ SemEval 2020
\citet{patwa2020sentimix} released 20K \textit{Hinglish} tweets, are annotated with word-level languages and sentence-level sentiment i.e. \textit{ positive, negative, neutral}. \citet{liu2020kk2018} achieved the SOTA (75\% f1 score) by fine-tuning a pre-trained XLM-R using adversarial training.  

\citet{DBLP:journals/corr/VaswaniSPUJGKP17} introduced \textit{Positional Encoding (PE)} for language modeling. PE serves as an added feature along with the word embeddings, providing both \textit{relative} and \textit{absolute} positional relations between a target word and its context words. %To the best of our knowledge, PE has not been explored before for a CM setting.

%\vspace{-2mm}
\subsection{Absolute Positional Encoding (APE)} \textbf{Sinusoidal PE:} -  A predefined sinusoidal vector $p_{i} \in R^d$ is assigned to each position \textit{i}. This $p_{i}$ is added to the word embedding $w_{i} \in R^d$ at position \textit{i}, and $w_{i} + p_{i}$ is used as input to the model. In this way, the Transformer can differentiate the words coming from different positions and assign each token position-dependent attention \citet{DBLP:journals/corr/VaswaniSPUJGKP17}. Sin/cos functions are used interchangeably to capture odd/even numbered positional words in a sequence - equation~\ref{eq:1}

%\vspace{-4.5mm}
\begin{equation}
\alpha_{ij} ^{abs} = \frac{1}{\sqrt{d}} \left ( \left ( w_i + p_i  \right) W^{Q,1}\right)  \left ( \left ( w_j + p_j  \right) W^{K,1}\right) ^ T
\label{eq:1}
\end{equation}

%\vspace{-3mm}
%\begin{equation}
%\centering
%PE_{(pos, 2i)} = sin(pos/1000^{2i/d_{model}})
%\end{equation}

%\vspace{-3mm}
%\begin{equation}
%\centering
%PE_{(pos, 2i+1)} = cos(pos/1000^{2i/d_{model}})
%\end{equation}
\textbf{Dynamic PE:} -  Instead of using periodical functions like $sin/cos$,  \citet{DBLP:conf/icml/LiuYDH20}, proposed to learn a dynamic function at every encoder layer that can represent the positional info. A function $\theta(i)$ is introduced which can learn positional info with gradient flow. - equation~\ref{eq:2}

%\textcolor{blue}{Improving upon sinusoidal PE, Dynamic PE learns $\theta(i)$ instead of a predefined $p_i$ to bring dynamic behavior to the model. And at each utterance, this learnable function $\theta(i)$ tries to learns best possible representation for positional information with gradient flow and $\theta(i)$ is added to the word embedding $w_i$. Now, taking $w_i + \theta(i)$ as the input to transformer model. }
%\textcolor{red}{we need to describe the following equation}  
%\vspace{-6mm}
\begin{equation}
    \alpha_{ij} = \frac{1}{\sqrt{d}} \left (  \left ( w_i + \theta \left (i \right) \right) W^{Q,1}\right)  \left ( \left( w_j + \theta \left (j \right) \right) W^{K,1}\right) ^ T
    \label{eq:2}
\end{equation}

%Learning PE is inductive (can handle sequences longer than those seen during training), Data-Driven (learnable from data) and Parameter Efficient (number of trainable parameters introduced by the encoding should be limited).

%\vspace{-4.5mm}

\subsection{Relative Positional Encoding (RPE)} 
%In APE $p_{i}$ is fixed set of numbers, whereas, in Dynamic PE $\theta(i)$ is a learnable function. 
\citet{DBLP:journals/corr/abs-1803-02155} introduced a learnable parameter $a_{j-i}^l$ which learns the positional representation of the relative position \textit{j-i} at encoder layer \textit{l}. This helps the model to capture relative word orders explicitly - equation~\ref{eq:3}

%In absolute PE, using different \textit{pi} for different position \textit{i} helps transformer distinguish words at different positions. But, the absolute PE is not effective to capture the relative word ordering. \citet{DBLP:journals/corr/abs-1803-02155} proposed a relative positional encoding as an inductive bias to help the learning of attention modules. \textcolor{blue}{In addition to APE, RPE introduces a learnable parameter $a_{j-i}^l$ which learns the positional representation of the relative position \textit{j-i} at encoder layer \textit{l}. This helps the model to capture relative word orders explicitly.}
%\textcolor{red}{what is the difference between Dynamic vs. relative is not coming out properly.} 

%\vspace{-5.5mm}
\begin{equation}
\alpha_{ij}^{rel} = \frac{1}{\sqrt{d}} \left ( \left (w_i \right) ^ l W^{Q,l}\right)  \left ( \left  (w_i \right) ^ l  W^{K,l} + a_{j-i}^l \right) ^ T
\label{eq:3}
\end{equation}
%\begin{equation}
%a_{ij}^K  = w_{clip(j-1, k)} ^ K, a_{ij}^V  = w_{clip(j-1, k)} ^ V
%\end{equation}
 
%\vspace{-3mm}
\section{Switching Point based Positional Encoding} 
%\section{PESTO - Relative Positional Encoding that learns to emphasis on Switching Points}
We introduce a novel, switching point based PE. Consider the \textit{Hinglish} sequence - \textit{gaaye}\textsubscript{HI} \textit{aur}\textsubscript{HI} \textit{dance}\textsubscript{EN} \textit{kare}\textsubscript{HI}. SP based indices (SPI) - i) We set the index to 0 whenever an SP occurs. Indexing would normally be = $\{0,1,2,3\}$, we change it to $\{0,1,0,0\}$. ii)  We consider Hindi as our base language and English as the mixed language. We set the index to 0 only when the shift is from base language (\textit{L1}) to the mixed language (\textit{L2}). So, the resultant index would be $\{0,1,0,1\}$.

\subsection{Switching Point based Dynamic PE (SPDPE) }
We introduce a function $S(l_i)$, which takes the word level language labels as input and returns SPI. Instead of passing an index directly as i to $\theta$, we use $\theta(s(l_i))$ to dynamically learn the PE based on SPI - equation ~\ref{eq:4}
%\textcolor{blue}{Instead of training by using sequential indexing like (0,1,2,3..), we train dynamic PE with SP based indices. We get our SP based indices using our function $S(l_i)$. This function takes the word level label and returns SP based indices. $S(l_i)$ checks for the SPs and it will set the index to zero whenever an SP occurs. This will give model some information about SPs and can learn based on it. Introduction of SP based indexing in positional encoding has shown 5\% improvement in F1 score. In the equation - \ref{eq:4}, we make use of our function S($l_i$). Instead of passing index directly as i to $\theta$, we use $\theta(s(l_i))$ to dynamically learn the positional encoding based on SP indices.}

% We train the dynamic PE with SP based indexes instead of usual indexing. Introduction of this gives an improvement of 5\% in weighted F1 score over other methods. 
%\textcolor{red}{not clear.}
%\vspace{-5mm}
\begin{equation}
\resizebox{.9\hsize}{!}{
$\alpha_{ij} = \frac{1}{\sqrt{d}} \left ( \left ( w_i + \theta \left (S(l_i) \right) \right)  W^{Q,1}\right) \\  \left ( \left (w_j + \theta \left (S(l_j) \right) \right)  W^{K,1} \right) ^ T$}   
\label{eq:4}
\end{equation}
%\vspace{-1mm}
\subsection{PESTO - Switching Point based Dynamic and Relative PE (SPDRPE)}
Here, in addition to the SPDPE, we use a learning parameter $a_{j-i}^l$, which encodes the relative position \textit{j-i} at the encoder layer \textit{l}. This encoding approach learns representations dynamically based on SPs along with the embedding $a_{j-i}^l$ so that it can also capture relative word orders (equation \ref{eq:5}). %This framework outperforms the SOTA \cite{liu2020kk2018}.

% To capture the relative word orders with the SP phenomena, we incorporate both SP based indexes and relative PE. Our models achieves outperforms the SOTA \cite{liu2020kk2018}. 
%\textcolor{red}{not clear.}
%\vspace{-6mm}
\begin{equation}
\resizebox{.9\hsize}{!}{
$\alpha_{ij} = \frac{1}{\sqrt{d}} \left ( \left ( w_i + \theta \left (S(l_i) \right) \right) ^ l W^{Q,l}\right)  \left ( \left (w_j + \theta \left (S(l_j) \right) \right) ^ l  W^{K,l} + a_{j-1}^l \right) ^ T$}
\label{eq:5}
\end{equation}

%\vspace{-4mm}

%We train FastText embeddings from scratch on our data. FastText embedding extracts local dependencies at word and sub-word level.  The switching point indices are given to the proposed 12 headed Multi Head Attention (12HA) transformer with relative and dynamic switching point based representation. The outputs of fastText and the relative and dynamic SP based representation are added and given to further layers in the encoder. 
% On top of the transformer encoder, a 1D CNN is used to get the representation of the entire sentence. We also obtain the sentence level embedding using tf-idf weighted average fasttext embedding. We concatenate the representations of the CNN and sentence level embedding and then pass it to a feed-forward network followed by softmax to predict the sentiment. The diagram of our model is given in Figure \ref{fig: Model}.

%\vspace{-3mm}
\section{Models}
\textbf{Baselines - Word2Vec, Multi Head Attention (MHA)}:
We choose Word2Vec as the baseline since it does not capture position info. We also choose attention mechanism, which is widely used to capture relational dependencies, to see its effects over SPs. We experiment with two lengths - i) Length 3 to capture the local window of dependency, whereas, ii) 12 to see whether it can learn anything from the whole sentence. 12 is the average length of sentences in our corpus.

\textbf{PESTO Overall Architecture:} 
The local dependencies from skipgram Word2Vec (trained from scratch) along with SPI obtained from SPDRPE are passed to a 12 headed transformer based encoder layer. %The output of this attention layer is passed to point-wise Feed forward network which encodes every word representation.
On top of the transformer, a 1D CNN is used to get the sentence level representation. We also obtain the sentence embedding using tf-idf weighted average of Word2Vec embeddings. Finally, we concatenate the representations of the CNN and the tf-idf sentence embedding and pass it to a dense layer which applies softmax to predict the sentiment. The architecture of PESTO is shown in Fig. \ref{fig: Model}. We train the entire model (2 encoder layers) from scratch, without using any pre-trained model.

%% file: 4_experiment_result.tex
%\vspace{-3mm}
\section{Results}
\begin{figure}[ht!]
\centering
%\subfigure[Word2Vec+LSTM]
%{\includegraphics[width=0.40\linewidth]{img/ft.png}}
\subfigure[Sinusoidal PE]
{\includegraphics[width=0.30\linewidth]{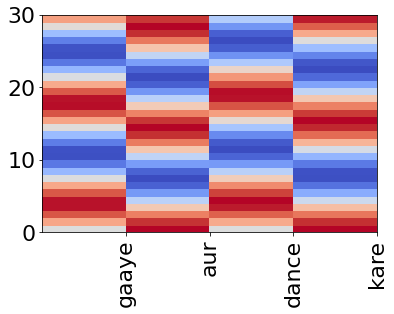}}
\subfigure[Dynamic PE]
{\includegraphics[width=0.30\linewidth]{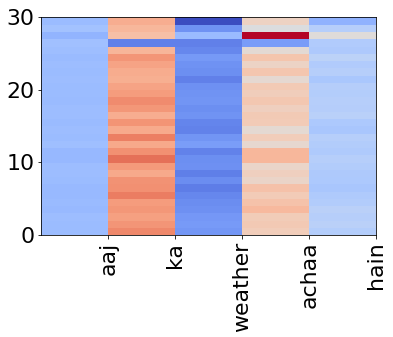}}
\subfigure[PESTO]
{\includegraphics[width=0.35\linewidth]{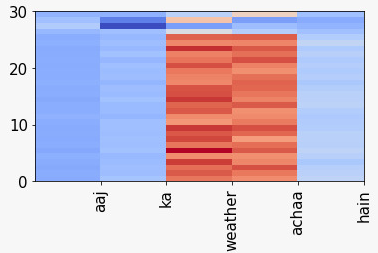}}
%\vspace{-5mm}
\caption{PESTO not only differentiates words coming from different positions, but also pays high attention to the SPs like \textit{weather}\textsubscript{EN} and \textit{achaa}\textsubscript{HI}.}
\label{fig: plot}
\end{figure}

PESTO achieves 75.56\% F1 score and outperforms SOTA (Tab. \ref{table:results}). The main reason for this is learning SP by aggregating both relative and dynamic PE with a variable length MHA framework. PESTO is able to generate more thrust to the switching point \textit{weather}\textsubscript{EN} \textit{achaa}\textsubscript{HI}  (Fig. \ref{fig: plot}). The experiments were conducted on google Colab. The code is available at \url{https://github.com/mohammedmohsinali/PESTO}.

\begin{figure}[ht!]
\centering
\includegraphics[width=0.6\linewidth]{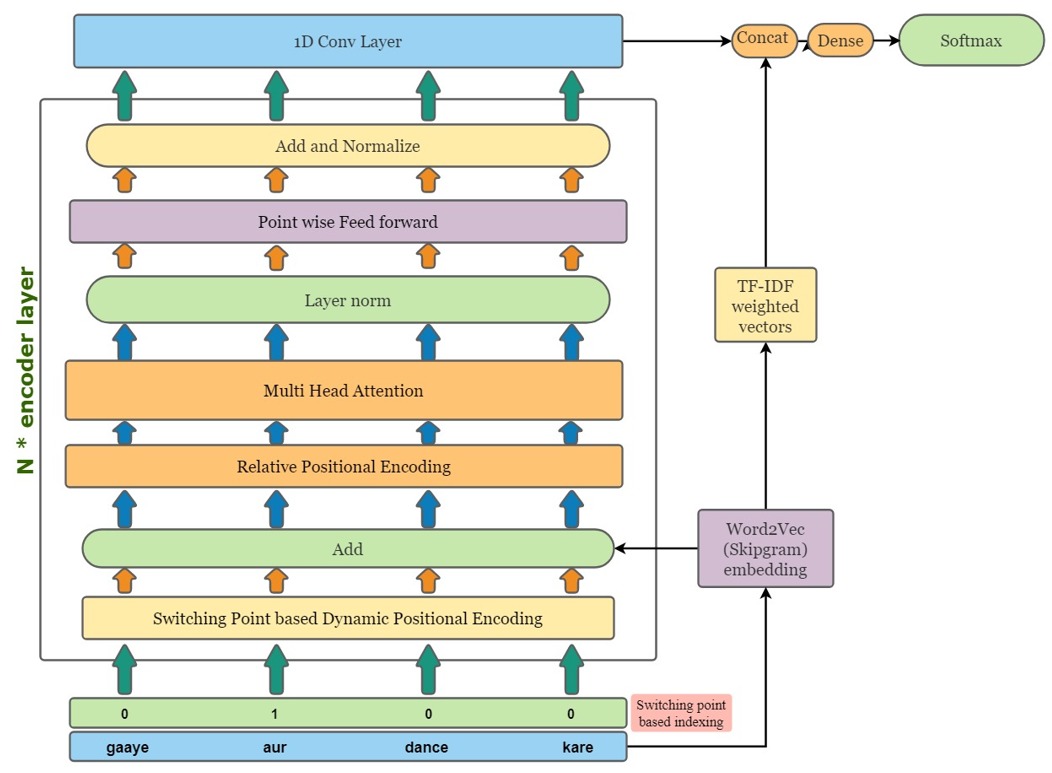}

\caption{PESTO - Proposed model for Relative Switching Point based Dynamic Positional Representation for CM text. }
\label{fig: Model}
\end{figure}

\begin{table}[ht!]
\centering
\resizebox{\columnwidth}{!}{%
\begin{tabular}{cccccc}
\hline
\multirow{2}{*}{Models} & \multicolumn{4}{c}{Positional Representation} & F1 (\%)\\ \cline{2-6} 
 & Sin/Cos & Index & SPI & Relative &  \\ \cline{1-6}
Word2Vec + LSTM & \xmark & \xmark & \xmark & \xmark & 56 \\
\hline
ELMO &  \xmark & \xmark  & \xmark & \xmark & 55 \\
BERT & \cmark & \xmark & \xmark & \xmark & 60 \\
3HA with Sinusoidal PE & \cmark & \xmark & \xmark & \xmark & 65 \\
3HA with Dynamic PE & \xmark & \cmark & \xmark & \xmark & 69.7 \\
\hline
12HA with Dynamic PE+RPE   & \xmark & \cmark & \xmark & \cmark & 73 \\
12HA with RPE & \xmark & \xmark & \xmark & \cmark & 73.4 \\
12HA with SPDPE & \xmark & \xmark & \cmark & \xmark & 73.52 \\
SOTA \cite{liu2020kk2018} & \cmark & \xmark & \xmark & \xmark & 75 \\
\textbf{PESTO (12HA with SPDRE)} & \xmark & \xmark & \cmark & \cmark &  \textbf{75.56} \\ \hline

\end{tabular}%
}

\caption{Results of various position sensitive experiments on CM text. It is evident from the results that positional encoding could be an effective way to train position sensitive language models for CM text.}
\label{table:results}
\end{table}

%% file: 5_conclusion.tex
\section{Conclusion}

In this paper we report initial experiments on \textit{Hinglish} sentiment analysis problem through the lens of language modeling. We argued SPs are the major bottleneck for CM. Our contribution could be seen as following - i) We introduce the idea of switching-point based positional encoding. i) We propose a relative switching point dynamic positional encoding technique named PESTO, which yields better results than SOTA.  iii) It is also noteworthy that our model - PESTO achieves SOTA results without any pre-trained heavy language model, whereas all the SOTA models in the SentiMix task used models like BERT, or XLNet.

%We observe that one of the biggest challenges of CM data is SP. Keeping this in mind, We propose an efficient framework to learn SP based relative and dynamic PE, which shows better results over other models. Dynamic positional and relative encoding not only help in differentiating the text but also help in learning a new representation that is based on SP.

%In this paper, our contribution is as follows :

%(i) we have collected over 80k hinglish twitter data and have trained word embeddings on over 100k hinglish corpus (combination of our collected twitter data, ICON-2016 dataset, Semeval-2020 dataset) and further evaluated the performance of these embeddings on similarity files (wordsim-353, simlex-999)

%(ii) For the problem of sentiment analysis, we propose an efficient neural architecture that can also learn based on switching points. We have designed a multi head attention model that can learn relative switching point based dynamic positional representations. We have also observed that introducing switching point based relative positioning has shown a significant improvement in performance.